\PassOptionsToPackage{usenames, dvipsnames}{color}
\documentclass[final]{cvpr}

\usepackage{times}
\usepackage{epsfig}
\usepackage{graphicx}
\usepackage{helvet}
\usepackage{booktabs}
\usepackage{courier}
\usepackage{amsmath}
\usepackage{algorithm}
\usepackage{algorithmic}
\usepackage{csquotes} 
\usepackage{color}
\usepackage{paralist}
\usepackage{amssymb}
\usepackage{indentfirst}
\usepackage{subfigure}
\usepackage{float}
\usepackage{multirow}
\usepackage{cite}
\usepackage{mathrsfs}
\usepackage[colorlinks, linkcolor=red,  anchorcolor=blue, citecolor=blue]{hyperref}
\usepackage{textcomp,booktabs}
\usepackage{amssymb}
\usepackage{pifont}
\newcommand{\cmark}{\ding{51}}%
\newcommand{\xmark}{\ding{55}}%

\usepackage{colortbl}
\definecolor{mygray}{gray}{.9}
\definecolor{mypink}{rgb}{.99,.91,.95}
\definecolor{mycyan}{cmyk}{.3,0,0,0}
\def\cvprPaperID{1435} 
\def\confYear{CVPR 2021}

\begin{document}

\title{Towards More Flexible and Accurate Object Tracking with Natural Language: Algorithms and Benchmark	}
\author{
Xiao Wang$^{1}$\thanks{The first two authors contribute equally to this work. Yaowei Wang is the corresponding author. Email: \{wangx03, shuxj, wangyw, tianyh\}@pcl.ac.cn, zhangzhipeng2017@ia.ac.cn, jiangbo@ahu.edu.cn, fengwu@ustc.edu.cn.}, Xiujun Shu$^{2,1*}$, Zhipeng Zhang$^{3}$, Bo Jiang$^{4}$, Yaowei Wang$^{1}$, Yonghong Tian$^{1,5}$, Feng Wu$^{1,6}$ \\ 
${^1}$Peng Cheng Laboratory, Shenzhen, China \\
${^2}$School of Electronic and Computer Engineering, Peking University, Shenzhen, China  \\
${^3}$NLPR, Institute of Automation, Chinese Academy of Sciences \\
${^4}$School of Computer Science and Technology, Anhui University, Hefei, China \\
${^5}$Department of Computer Science and Technology, Peking University, Beijing, China  \\
${^6}$University of Science and Technology of China, Hefei, China \\
\url{https://sites.google.com/view/langtrackbenchmark/}
}

\maketitle

\begin{abstract}
Tracking by natural language specification is a new rising research topic that aims at locating the target object in the video sequence based on its language description. Compared with traditional bounding box (BBox) based tracking, this setting guides object tracking with high-level semantic information, addresses the ambiguity of BBox, and links local and global search organically together. Those benefits may bring more flexible, robust and accurate tracking performance in practical scenarios. However, existing natural language initialized trackers are developed and compared on benchmark datasets proposed for tracking-by-BBox, which can't reflect the true power of tracking-by-language. In this work, we propose a new benchmark specifically dedicated to the tracking-by-language, including a large scale dataset, strong and diverse baseline methods. Specifically, we collect 2k video sequences (contains a total of 1,244,340 frames, 663 words) and split 1300/700 for the train/testing respectively. We densely annotate one sentence in English and corresponding bounding boxes of the target object for each video. We also introduce two new challenges into TNL2K for the object tracking task, i.e., adversarial samples and modality switch. A strong baseline method based on an adaptive local-global-search scheme is proposed for future works to compare. We believe this benchmark will greatly boost related researches on natural language guided tracking. 
\end{abstract}


\section{Introduction}
\begin{figure} 
\center
\includegraphics[width=3.3in]{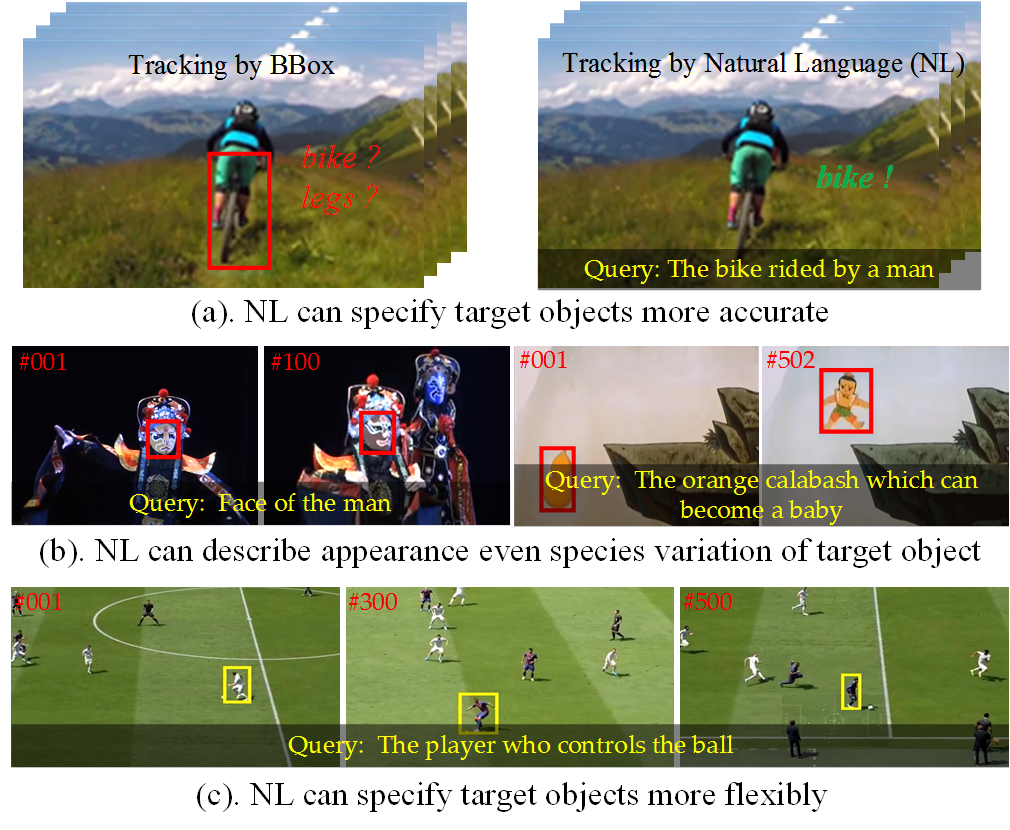}
\caption{Comparison between the task of tracking-by-BBox and tracking-by-language. We can find that tracking-by-NL can specify target object more accurate and flexibly, and is also good at describing the appearance/species variation.}
\label{motivation}
\end{figure}

Single object tracking is one of the most important tasks in computer vision and it has been widely used in many applications such as video surveillance, robotics, and autonomous vehicles. Usually, they initialize the target object in the first frame with a bounding box (BBox), as shown in Fig. \ref{motivation} (a), and adjust the BBox along with the movement of the target object. Most of the existing single object trackers \cite{Wang_2018_CVPR, Henriques2015High, hare2016struck, held2016GOTURN, Yun2017ADNet, wang2019GANTrack} are developed based on this setting\footnote{\url{https://github.com/wangxiao5791509/Single_Object_Tracking_Paper_List}}, and many benchmark datasets \cite{Wu2013Online, Liang2015Encoding, wu2015object, mueller2016benchmarkuav20l, huang2019got10k, fan2019lasot, valmadre2018longOxUVA} are proposed for this task.

Although these trackers have been adopted in many applications, however, the setting of tracking-by-BBox still suffers from the following issues. 
(1) The target object in the first frame with a BBox is inconvenient to initialize in practical scenarios. In another word, the initialization limits the wide applications of existing BBox initialized trackers. 
(2) The initialized BBox may be not optimal for the representation of target object which may lead to ambiguity. As shown in Fig. \ref{motivation} (a), the tracker may be confused to track the \emph{bike} or \emph{lower body} of the pedestrian. Similar views can also be found in \cite{li2017tracking, wang2018describe, feng2019robust, yang2019grounding}. 
(3) Current BBox-based trackers may perform poorly when facing abrupt appearance variation of the target object, like \emph{face/cloth changing or species variation} in Fig. \ref{motivation} (b). Because the appearance feature initialized in the first frame and the object in the tracking procedure are vastly different. Only one sample initialized in the first frame is not enough to handle these challenging scenarios. 
These observations all inspire us to begin to think about \emph{how can we conduct tracking in a more applicable and accurate way?}

Recently, some researchers attempt to introduce the natural language description instead of the BBox for tracking \cite{li2017tracking, wang2018describe, feng2019robust, yang2019grounding}, termed tracking by natural language. This setting allows for a new type of human-machine interaction in object tracking. For example, it can enhance existing BBox based trackers by helping them against model drift, or simultaneous multiple-video tracking as noted in \cite{li2017tracking}. More importantly, natural language is more convenient and intuitive to express for human beings compared with BBox. It can provide a more precise expression of the target object from spatial location to high-level semantic information like \emph{attributes, category, shape, properties}, and \emph{structural relationship} with other objects, etc. This information will be beneficial to address the ambiguity issue of BBox and the vast appearance variation of the target object. Meanwhile, the language can also specify target objects more flexibly, for example, ``\emph{The player who controls the ball}" in Fig. \ref{motivation} (c). The intelligent tracker should focus on target players even the ball passed to different persons, without having to re-initialize the target person like the standard setup of visual tracking. However, this research topic has received far less attention than standard target tracking. Only a few works \cite{li2017tracking, wang2018describe, feng2020langTrackwacv, feng2019robust, yang2019grounding} are developed and compared on tracking benchmark datasets specially designed for BBox based tracking. These benchmarks may fail to nicely reflect the true power of tracking-by-language, and this inspires us to design a new and large-scale benchmark for this task.

In this work, we collect a large-scale dataset that contains $2,000$ video sequences, named TNL2K. These videos are collected from YouTube\footnote{\url{https://www.youtube.com/}}, surveillance cameras, and mobile. For each video, we densely annotate the location information of the target object for each frame and one sentence in English for the whole video. Specifically, we describe the category, shape, attributes, properties, and spatial location of the target object which will provide rich fine-grained \emph{appearance information} and high-level \emph{semantic information} for tracking. We select $1,300$ videos for training and the rest $700$ videos for evaluation. Our videos also reflect two attributes for the tracking task, i.e., the adversarial samples and modality switch between RGB and thermal data. To provide a baseline method for other researchers to compare, we design a simple but strong algorithm based on an adaptive local-global-search scheme. Specifically, three kinds of baseline results are provided, i.e., Tracking-by-BBox, Tracking-by-Language, Tracking-by-BBox and Language.

The contributions of this paper can be summarized in the following three aspects: 

$\bullet$ We propose the TNL2K dataset for the natural language-based tracking which consists of $2,000$ video sequences. It aims at offering a dedicated platform for the development and assessment of natural language-based tracking algorithms. 

$\bullet$  We propose a simple but strong baseline approach (termed AdaSwitcher) for future works to compare, which can switch between the local tracking algorithm and global grounding module adaptively.  

$\bullet$  To provide extensive baselines for the comparison on TNL2K dataset, we also evaluate more than 40 representative BBox-based trackers and analyze their performance using different evaluation metrics.

\section{Related Work} 

\textbf{Tracking by Bounding Box }
The standard trackers begin their tracking procedure based on an initialized BBox in the first frame, including classification based \cite{Nam2015Learning, Jung_2018_ECCV, Park_2018_ECCV, han2017branchout}, Siamese network based \cite{Wang_2018_CVPR, li2018siamrpn++, xu2020siamfc++, danelljan2019atom, danelljan2020PRDiMP}, correlation filter based \cite{lukezic2017discriminative, Danelljan2016Beyond, Henriques2015High, Danelljan2015Learning}, and regression-based \cite{held2016GOTURN}. Inspired by the success of neural networks on image classification, most of the recent trackers are developed based on deep learning. Specifically, the Siamese network based trackers achieve state-of-the-art performance on multiple tracking benchmarks. Previous Siamese trackers simply measure the similarity between the static target template with extracted proposals and treat the best-scored proposal as their tracking results. Recently, some researchers begin to collect the tracking results which can be used to dynamically update the target template and attain better results \cite{Yang_2018_ECCV, zhang2019UpdateNet}. In addition to learn powerful feature representation and conduct a local search for tracking, some trackers attempt to achieve robust tracking by global search \cite{Goutam2020KYS, voigtlaender2020siamRCNN, yan2019skimming, wang2019GANTrack, huang2019globaltrack, deepMTA}.  For more related works on standard visual tracking, please check the following survey papers \cite{li2013survey, Smeulders2014Visual, Yilmaz2006Object, marvasti2019TrackSurvey, li2018TrackSurvey}.

\textbf{Tracking by Natural Language } 
Due to it is a new rising topic, only a few algorithms are developed and the authors of \cite{li2017tracking} first validated the effectiveness of natural language for the tracking task by designing three modules (i.e. Lingual Specification Only; Lingual First, then Visual Specification; Lingual and Visual Specification). Wang \cite{wang2018describe} and Feng \cite{feng2019robust} also propose to use the language information to generate global proposals for tracking. Yang et al. propose the GTI \cite{yang2019grounding} which decomposes the tracking problem into three sub-tasks, i.e., grounding, tracking and integration, and these modules operate simultaneously and predict the box sequence frame-by-frame. These methods are evaluated on datasets specifically designed for tracking-by-BBox which may fail to reflect the feature of tracking-by-language. To the best of our knowledge, there is still no public benchmark specifically dedicated to the tracking-by-language task. We believe our benchmark will greatly boost the researches on natural language related object tracking.

\textbf{Benchmarks for Tracking }
Existing benchmarks for visual tracking can be concluded into two main categories according to whether contains training data. As shown in Table \ref{benchmarkList}, previous benchmarks \cite{Liang2015Encoding, Wu2013Online, wu2015object, VOT_TPAMI, nus_pro, li2017visualuav, li2017visualuav, kiani2017Nfs} provide test videos only before deep trackers occurred. It is worthy to note that OTB-2013 \cite{Wu2013Online} and OTB-2015 \cite{wu2015object} are the first public benchmarks for visual tracking which contain 50 and 100 video sequences, respectively. In the deep learning era, several large scale tracking benchmarks are proposed for the training of deep trackers. For example, GOT-10k \cite{huang2019got10k} contains $10,000$ videos which can be categorized into 563 classes. TrackingNet \cite{muller2018trackingnet} is a subset (31K sequences selected) of video object detection benchmark YT-BB \cite{real2017YTBB} and the ground truth is manually labeled at 1 FPS. OxUvA \cite{valmadre2018longOxUVA} and LaSOT \cite{fan2019lasot} are two long-term tracking benchmark which consists of 366 and 1400 video sequences respectively.

The aforementioned tracking benchmarks are all mainly designed for tracking by BBox, although the LaSOT indeed provides the language specification of the target object. However, they only describe the appearance of the target object but ignore the relative location which may limit the integration of natural language. In another word, their benchmark is suitable for natural language assisted tracking but is not for the task of language initialized tracking. Another issue of the existing benchmark is that these videos do not contain videos with significant appearance variations, such as clothing change for a pedestrian. This also limits the application of existing trackers in practical scenarios. Besides, these benchmarks also ignore the adversarial samples which limit the development of adversarial learning-based trackers \cite{wiyatno2019TrackAttack, jia2020TrackAttack, liang2020TrackAttack, yan2020TrackAttack}. By contrast, our proposed TNL2K is specifically designed for tracking by natural language specification and contains multiple videos with significant appearance variation and adversarial samples. It also contains natural videos, animation videos, infrared videos, virtual game videos, which are suitable for the evaluation of domain adaptation of current trackers. We also provide baseline results of three kinds of settings which will be beneficial for future trackers to compare.

\section{Tracking by Natural Language} 


\subsection{TNL2K Dataset} 

\begin{figure}[!htb]
\center
\includegraphics[width=3.3in]{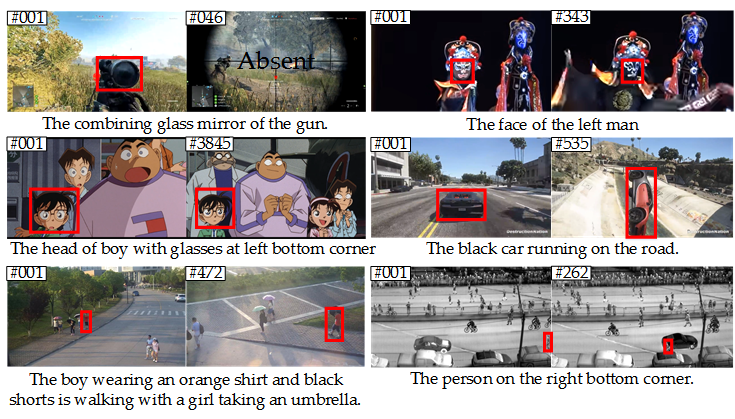}
\caption{Example sequences and annotations in TNL2K dataset. }
\label{videoSample}
\end{figure}

\textbf{Data Collection and Annotation}
The proposed TNL2K dataset contains $2,000$ video sequences, and most of them are downloaded and clipped from YouTube, intelligent surveillance cameras, and mobile phones. We invite seven people for the annotation of these videos. Specifically, we annotate one sentence in English for each video and also one bounding box for each frame in this video. The left corner point ($x_1, y_1$), width $w$ and height $h$ of the target's bounding box are used as the ground truth, i.e., [$x_1, y_1, w, h$]. The annotated natural language description indicates the \emph{spatial position}, \emph{relative location with other objects}, \emph{attribute}, \emph{category} and \emph{property} of target object in the first frame. We also annotate the \emph{absent} label for each frame to enrich the information that is available for more accurate tracking. To construct a rich and heterogeneous benchmark, we also borrow some thermal videos from existing datasets \cite{li2019rgbt234, liu2019ptbtracking} and re-annotate the target object we want to track if necessary. Example sequences and annotations are illustrated in Fig. \ref{videoSample}.

\begin{table*}[htp]
\center
\scriptsize 
\caption{Comparison of current datasets for object tracking. $\#$ denotes the number of corresponding item. Lang-A and Lang-I denote the dataset can be used for language assisted and initialized tracking task. SAV denotes the dataset contains many videos with significant appearance variation. Adv means the dataset contains adversarial samples (i.e., malicious attacks). DA is short for domain adaptation.} \label{benchmarkList}
\begin{tabular}{l|ccccccccccccccc}
\hline \toprule [0.8 pt]
\textbf{Datasets}    &\textbf{\#Videos}  &\textbf{\#Min} &\textbf{\#Mean} &\textbf{\#Max} &\textbf{\#Total} &\textbf{\#FR}  &\textbf{\#Attributes} &\textbf{Aim} &\textbf{Absent} &\textbf{Lang-A} &\textbf{Lang-I} & \textbf{SAV}      &\textbf{Adv}   &\textbf{DA} \\ 
\hline
\textbf{OTB50 \cite{Wu2013Online}}   	    &51       &71 	&578    &3,872    &29K    &30 fps      &11    &Eval      &       &     &           &             &          &           \\
\textbf{OTB100 \cite{wu2015object}}   	&100     &71  	&590    &3,872    &59K    &30 fps     &11    &Eval    &      &         &           &          &         &            \\
\textbf{TC-128 \cite{Liang2015Encoding}}   	    &128     &71  	&429    &3,872    &55K    &30 fps       &11    &Eval     &    &      &            &         &            &             \\
\textbf{VOT-2017 \cite{VOT_TPAMI}}   &60      &41  	&356    &1,500    &21K    &30 fps     &-    &Eval    &       &     &     &                 &             &          \\
\textbf{NUS-PRO \cite{nus_pro}}   	&365     &146  	&371    &5040    &135K    &30 fps      &-    &Eval    &       &      &          &            &          &            \\
\textbf{UAV123 \cite{mueller2016benchmarkuav20l}}   	&123     &109  	&915    &3085    &113K    &30 fps         &12    &Eval    &       &      &     &            &               &            \\
\textbf{UAV20L \cite{mueller2016benchmarkuav20l}}   	&20     &1717  	&2934    &5527    &59K    &30 fps          &12    		&Eval    &       &       &  &           &              &                \\
\textbf{NfS \cite{kiani2017Nfs}}   	            &100     &169  	&3830    &20665    &383K    &240 fps           &9    &Eval    &       &         &      &            &          &             \\
\hline
\textbf{TrackingNet \cite{muller2018trackingnet}} 	&30,643     & - 		&480   	 & -   		& 14.43M   	&30 fps           &15    &Train/Eval    &         &     &       &           &           &              \\
\textbf{OxUvA \cite{valmadre2018longOxUVA}}   	    		&366     	  & 900 	&4260   &37440   &1.55M  		&30 fps           &6    &Train/Eval     &         &    &         &            &           &           \\
\textbf{GOT-10k \cite{huang2019got10k}}   			&10,000    &29  		&149    &1,418     &1.5M    		&10 fps    		&6    &Train/Eval   &\cmark    &    &       &             &           &           \\ 
\textbf{LaSOT \cite{fan2019lasot}}   	    				&1,400      &1000		&2506   &11397    &3.52M      &30 fps         	&14    &Train/Eval  &\cmark           &\cmark    &          &       &        &            \\
\hline
\textbf{TNL2K (Ours)}  &2,000     &21  	&622    &18488    &1.24M    &30 fps       &17   &Train/Eval          &\cmark   &\cmark    &\cmark           &\cmark      &\cmark         &\cmark          \\
\hline \toprule [0.8 pt]
\end{tabular}
\end{table*}

\begin{figure}[!htb]
\center
\includegraphics[width=3.3in]{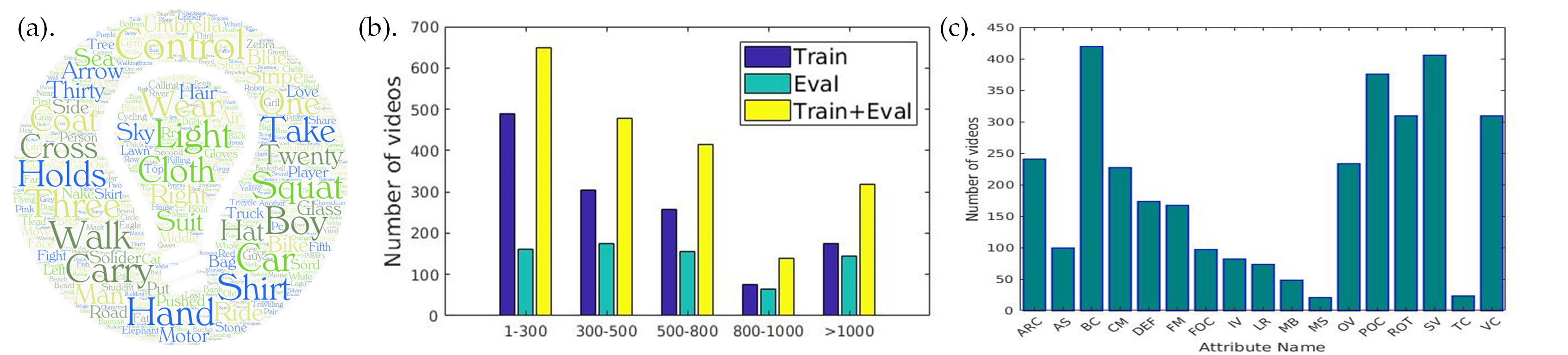}
\caption{ (a) Some words in our language description; (b, c) Distribution of sequences in each attribute and length in our TNL2K. Best viewed by zooming in.}
\label{dataAnalysis}
\end{figure}

\textbf{Attribute Definition}
Following popular tracking benchmarks \cite{fan2019lasot, huang2019got10k, wu2015object}, we also define multiple attributes of each video sequence for the evaluation under each challenging factors. As shown in Table \ref{AttributeList}, our proposed TNL2K dataset has the following 17 attributes: CM (Camera Motion), ROT (Rotate Of Target), DEF (DEFormation), FOC (Fully OCcluded), IV (Illumination Variation), OV (Out of View), POC (Partially OCcluded), VC (Viewpoint Change), SV (Scale Variation), BC (Background Clutter), MB (Motion Blur), ARC (Aspect Ratio Change), LR (Low Resolution), FM (Fast Motion), AS (Adversarial Sample), TC (Thermal Crossover), MS (Modality Switch). It is worthy to note that our dataset contains some thermal videos with challenging factors like TC (target object shares similar intensity with background), MS (the video contains both thermal and RGB images). To provide a good platform for the study of adversarial attack and defense of neural network for tracking, we also generate 100 videos contain adversarial samples as part of the testing subset using attack toolkit \cite{jia2020TrackAttack}. Therefore, these videos contain additional challenging factor, i.e., AS (influence of Adversarial Samples). It is worthy to note that the AS and MS are two new attributes for tracking community first proposed in this work. A more detailed distribution of each challenge is shown in Fig. \ref{dataAnalysis} (c).

\begin{table}[htp!]
\center
\scriptsize 
\caption{Description of 17 attributes in our TNL2K dataset.} \label{AttributeList}
\begin{tabular}{l|lcccccccccccccc}
\hline \toprule [0.8 pt]
\textbf{Attributes}    &\textbf{Definition}  \\ 
\hline
\textbf{01. CM}   	    	&Abrupt motion of the camera \\
\textbf{02. ROT}   	    &Target object rotates in the video \\
\textbf{03. DEF}   	    &The target is deformable \\
\textbf{04. FOC}   	    &Target is fully occluded \\
\textbf{05. IV}   	    	&Illumination variation \\ 
\textbf{06. OV}   	    	&The target completely leaves the video sequence \\ 
\textbf{07. POC}   	    &Partially occluded  \\
\textbf{08. VC}   	    	&Viewpoint change  \\
\textbf{09. SV}   	    	&Scale variation  \\
\textbf{10. BC}   	    	&Background clutter  \\
\textbf{11. MB}   	    	&Motion blur  \\
\textbf{12. ARC}   	    &The ratio of bounding box aspect ratio is outside the range [0.5, 2]   \\
\textbf{13. LR}   	    	&Low resolution  \\
\textbf{14. FM}   	    	&The motion of the target is larger than the size of its bounding box  \\
\textbf{15. AS}   	    	&Influence of adversarial samples  \\
\textbf{16. TC}			&Two targets with similar intensity cross each other \\ 
\textbf{17. MS}			&Video contain both color and thermal images \\ 
\hline \toprule [0.8 pt]
\end{tabular}
\end{table}

\textbf{Statistical Analysis}
Our proposed TNL2K contains 663 English words and focuses on expressing the attributes, spatial location of target objects, as shown in Fig. \ref{dataAnalysis} (a). For the distribution of length of all videos, we can see from Fig. \ref{dataAnalysis} (b) that the TNL2K contains [648, 479, 415, 139, 319] videos for the category of 1-300, 300-500, 500-800, 800-1000, and larger than 1000. More details, the number of these five segments for train and evaluation set are [488, 304, 258, 75, 175] and [160, 175, 157, 64, 144] respectively. We can find that our test set contains 144 long-term videos (larger than 1000 frames for each video) which will be suitable for the evaluation of long-term trackers. From Fig. \ref{dataAnalysis} (c), we can find that our TNL2K contains many videos with challenging attributes like \emph{background clutter}, \emph{scale variation}, \emph{view change},  \emph{partially occlusion}, \emph{out-of-view} and \emph{rotate}. The videos with these challenging factors will provide a good platform for the evaluation of current trackers. 

\subsection{Our Proposed Approach}  

\begin{figure}
\center
\includegraphics[width=3.3in]{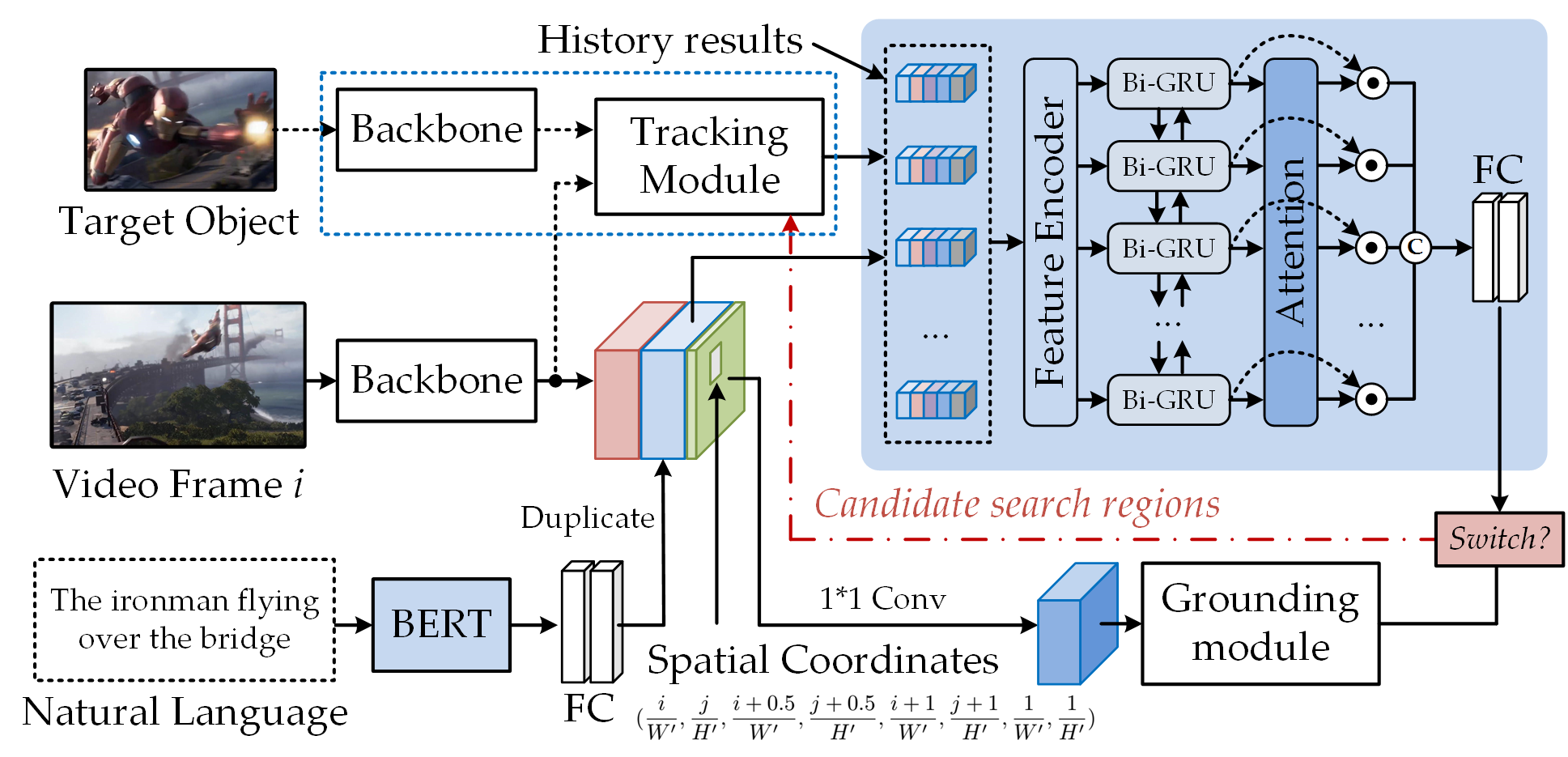}
\caption{An overview of our proposed adaptive tracking and grounding switch framework. AdaSwitcher is highlighted in blue.}  
\label{pipeline}
\end{figure}

In this paper, we propose the adaptive tracking and grounding switch framework for tracking by natural language specification, as shown in Fig. \ref{pipeline}. We will first introduce the visual grounding and visual tracking module, then, we will focus on our AdaSwitcher module.

\textbf{Visual Grounding Module}  
In the tracking by natural language task, we need to first locate the target object only depends on the language description $S = [w_1, w_2, ... , w_T]$. It is a standard visual grounding task and we follow the algorithm \cite{yang2019fastgrounding} proposed by Yang et al. due to its good performance and efficiency.

As shown in Fig. \ref{pipeline}, the visual grounding module takes video frame and natural language description as input. We use the backbone CNN to obtain the deep feature representation of the $i$-th video frame $F_i$. For the natural language, we first embed the words into feature representations $E = [e_1, e_2, ... , e_T]$ using a pre-trained BERT \cite{devlin2019bert} which is a widely used word embedding model in natural language related tasks. Then, this feature is fed into two fully connected layers for further fine tuning. Following \cite{yang2019fastgrounding}, we also duplicate this feature vector into feature maps and concatenate them with visual features of video frame. Another important information for visual grounding is the spatial coordinates encoding due to the spatial configurations are usually adopted to refer to target object. Therefore, the spatial feature for each position is also explicitly encoded in this work by following \cite{yang2019fastgrounding}.


The visual feature maps of global frame, duplicated language feature, and the spatial coordinates are concatenated together and fed into convolutional layers with kernel size $1 \times 1$ for information fusion. The output feature map is then sent into the grounding module, which will output the predicted location of target object. We treat such visual grounding as a global search procedure for tracking by natural language, which plays an important role at the beginning of the video and when we need to re-detection the target object in tracking procedure. The integration of visual grounding and tracker SiamRPN++ \cite{li2018siamrpn++} is termed Ours-I in Table \ref{Benchmarkresults}. Besides, we also explore the target-aware attention (termed TANet) proposed in \cite{wang2018describe, wang2019GANTrack}, i.e., Ours-II in Table \ref{Benchmarkresults}. The TANet takes the feature maps of target object and video image as input, and output corresponding global attention using de-convolutional network which can be used for search target object from global view. We refer the readers to check \cite{wang2018describe, wang2019GANTrack} for further understanding of this module.

\textbf{Visual Tracking Module} 
Aforementioned visual grounding can help detect the target object at the beginning, however, only grounding is not enough for high performance tracking, since it is easily influenced by background clutter. In this work, we initialize a  visual tracker for target object location in a local search manner based on the predicted bounding box from visual grounding in the first frame. The SiamRPN++ \cite{li2018siamrpn++} is adopted in our experiments due to its good performance.

\begin{figure} 
\center
\includegraphics[width=3.3in]{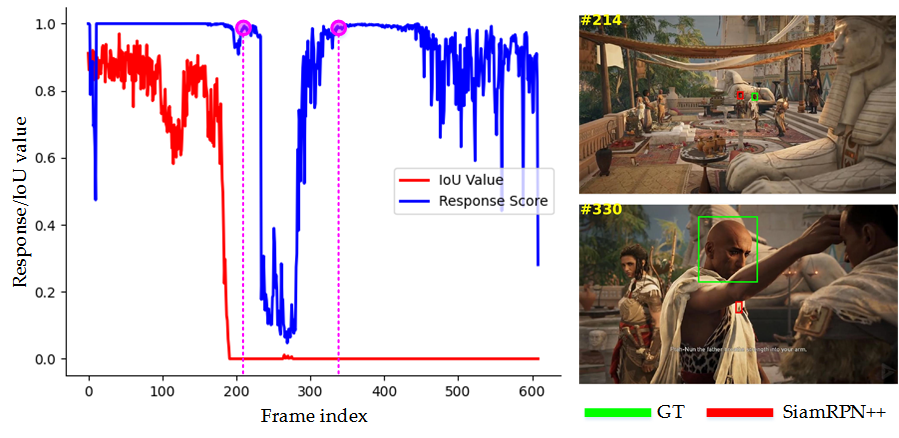}
\caption{Illustration of current trackers with high response score but low IoU values (take the SiamRPN++ \cite{li2018siamrpn++} as an example).}   
\label{scoreIoUVis}
\end{figure}

\textbf{AdaSwitcher Module} 
Given the visual grounding and visual tracking module, we can capture the target object from global and local view, respectively. One thorny issue that still exists is when we use visual grounding for global search (or visual tracking for local search). One intuitive approach is to conduct such switch based on the confidence of tracker, however, the confidence score is not always reliable especially in the challenging scenarios. For example, as shown in Fig. \ref{scoreIoUVis}, the confidence score is very high (larger than $0.9$) in some frames, but the model actually locates wrong object. Inspired by anomaly detection (also called outlier detection) whose target is the identification of rare items, events or observations which raise suspicions by differing significantly from the majority of the data. In this work, we take the failure of visual tracking as a kind of anomaly detection and propose a novel AdaSwitcher module to detect such failure. Once the anomaly is detected (the prediction from AdaSwitcher is larger than a pre-defined threshold), we can switch the candidate search regions from visual tracking to visual grounding for more robust and accurate tracking.

In this paper, \emph{confidence score} (1-D), \emph{BBox} (4-D), \emph{result image} (($30*30*3$)-D), \emph{response map} (($23*23$)-D) and \emph{language embedding} (512-D) are exploited in this work as the input of our AdaSwitcher. This information can be collected from visual tracker easily for each frame. And the historical information of past video frames can also contribute to current anomaly detection. Assume we use the history of past $N$ frames, then, the dimension of these input are $N \times 1$, $N \times 4$, $N \times (23*23)$, $N \times (30*30*3)$, and $N \times 512$, respectively. We use multiple parallel fully connected layers to encode this information and embed them into fixed feature vectors, specifically, we have $F = [F_s, F_b, F_{img}, F_{map}, F_{emb}]$, whose dimension are $N \times 10$, $N \times 10$, $N \times 512$, $N \times 512$, and $N \times 512$, respectively. Then, these features are concatenated and fed into a bi-directional GRUs \cite{chung2014GRUs} to learn the temporal information.

Inspired by the fact that various frames may contribute differently, we introduce attention mechanism to encode the inputs differently. The attention weights $\alpha_i (i=1, ... , N)$ can be obtained by the multilayer perceptron (MLP): 
\begin{equation}
\small 
\label{frameAttWeights} 
\{\alpha_1, \alpha_2, ... , \alpha_N\} = MLP([F_s, F_b, F_{img}, F_{map}, F_{emb}])
\end{equation}
where $[ , ]$ denotes concatenate operation. The attention weights $\alpha_i (i=1, ... , N)$ are stacked into feature vectors $\hat{\alpha_i} (i=1, ... , N)$ which have same dimension with feature representation $F^i (i=1, ... , N)$ of each frame $i$. Therefore, the attended feature representations can be obtained by:  
\begin{equation}
\small 
\label{frameAttention} 
[ \bar{F^1}, \bar{F^2}, ... , \bar{F^N} ] = [ \hat{\alpha_1} * F^1,  \hat{\alpha_2}*F^2, ... , \hat{\alpha_N} * F^N] 
\end{equation}
After that, two fully connected layers are used to determine whether we should switch the candidate search regions from current tracking result to grounding result.

\subsection{Implementation Details} 
\textbf{Training Phase} 
In our experiments, we directly use the pre-trained weights of baseline tracker for visual tracking. For the visual grounding module, we train it on the training subset of our TNL2K dataset which contains $1,300$ video sequences  for 40 epochs. The initial learning rate is 1e-4, batchsize is 5. The YOLO loss function is used for this network by following \cite{redmon2018yolov3, yang2019fastgrounding}. For the AdaSwitcher, we first collect the training data by running the baseline tracker on the training subset of our TNL2K dataset. In this process, we treat the video clips whose average IoU (Intersection over Union) score larger than 0.7 as the positive data, and less than 0.5 as the negative data. For the data with average IoU score range from 0.5 to 0.7, we directly discard them due to it may bring confusion to our model. Similar operations can also be found in \cite{Jung_2018_ECCV}. The learning rate is 1e-5, batchsize is 1, the Adagrad \cite{duchi2011adagrad} is adopted as optimizer and trained for totally 30 epochs. We consider the switch between visual tracking and grounding as a binary classification problem, therefore, the BCE loss function is selected for the training of AdaSwitcher.

\textbf{Inference Phase} 
In this benchmark, three kinds of baseline methods are studied: 1). \emph{Tracking by Natural Language only:} In this setting, only the natural language is provided for tracking, we need to first locate the target object using visual grounding module. Then, we can conduct adaptive tracking (SiamRPN++ \cite{li2018siamrpn++} used in this setting) and grounding for high performance object localization. 2). \emph{Tracking by Natural Language and BBox:} We take the natural language as an external modality and conduct robust tracking based on both language and BBox. SiamRPN++ \cite{li2018siamrpn++} and TANet \cite{wang2019GANTrack} are used in this setting. 3). \emph{Tracking by BBox only:} To construct a comprehensive benchmark, we also provide baseline results for tracking by BBox only, i.e., the standard setting of visual object tracking. All the evaluated trackers can be found in our supplementary materials.

\section{Experiments}

\subsection{Datasets and Evaluation Protocols} 
In our experiments, the OTB-Lang \cite{wu2015object, li2017tracking}, LaSOT \cite{fan2019lasot} and our proposed TNL2K dataset are used for the evaluation. The OTB-lang contains 99 videos released from \cite{wu2015object}, then, the natural language specification is provided by Li et al. \cite{li2017tracking}. The LaSOT is a recently released long-term tracking dataset that provides both bounding box and natural language annotations. The test subset of LaSOT contains 280 video sequences.

Two popular metrics are adopted for the evaluation of tracking performance, including \textbf{Precision Plot} and \textbf{Success Plot}. Specifically, Precision Plot illustrates the percentage of frames where the center location error between the object location and ground truth is smaller than a pre-defined threshold (20-pixels threshold is usually adopted). Success Plot demonstrates the percentage of frames the IoU of the predicted and the ground truth bounding boxes is higher than a given ratio. The evaluation toolkit of this paper can be found at: \url{https://github.com/wangxiao5791509/TNL2K_evaluation_toolkit}.

\subsection{Benchmark Results} 

\textbf{Results of Tracking by Natural Language Only}
As shown in Table \ref{Benchmarkresults}, Li et al. \cite{li2017tracking} attain 0.29$|$0.25 on the OTB-Lang dataset, while Feng et al. achieve 0.56$|$0.54 and 0.78$|$0.54 in  \cite{feng2020langTrackwacv} and \cite{feng2019robust} respectively. When we take the result of visual grounding in the first frame as the initialized bbox of visual tracker SiamRPN++, we achieve 0.24$|$0.19 on the OTB-Lang dataset. On the LaSOT and TNL2K dataset, we attain 0.49$|$0.51 and 0.06$|$0.11$|$0.11 respectively. We can find that our method is comparable with Li et al. on the OTB-Lang dataset. On the larger dataset LaSOT, we attain better results than Feng et al. \cite{feng2020langTrackwacv}. These experimental results demonstrate that our baseline method can also achieve good performance on existing LaSOT and our proposed TNL2K dataset.

\begin{figure} 
\center
\includegraphics[width=3.3in]{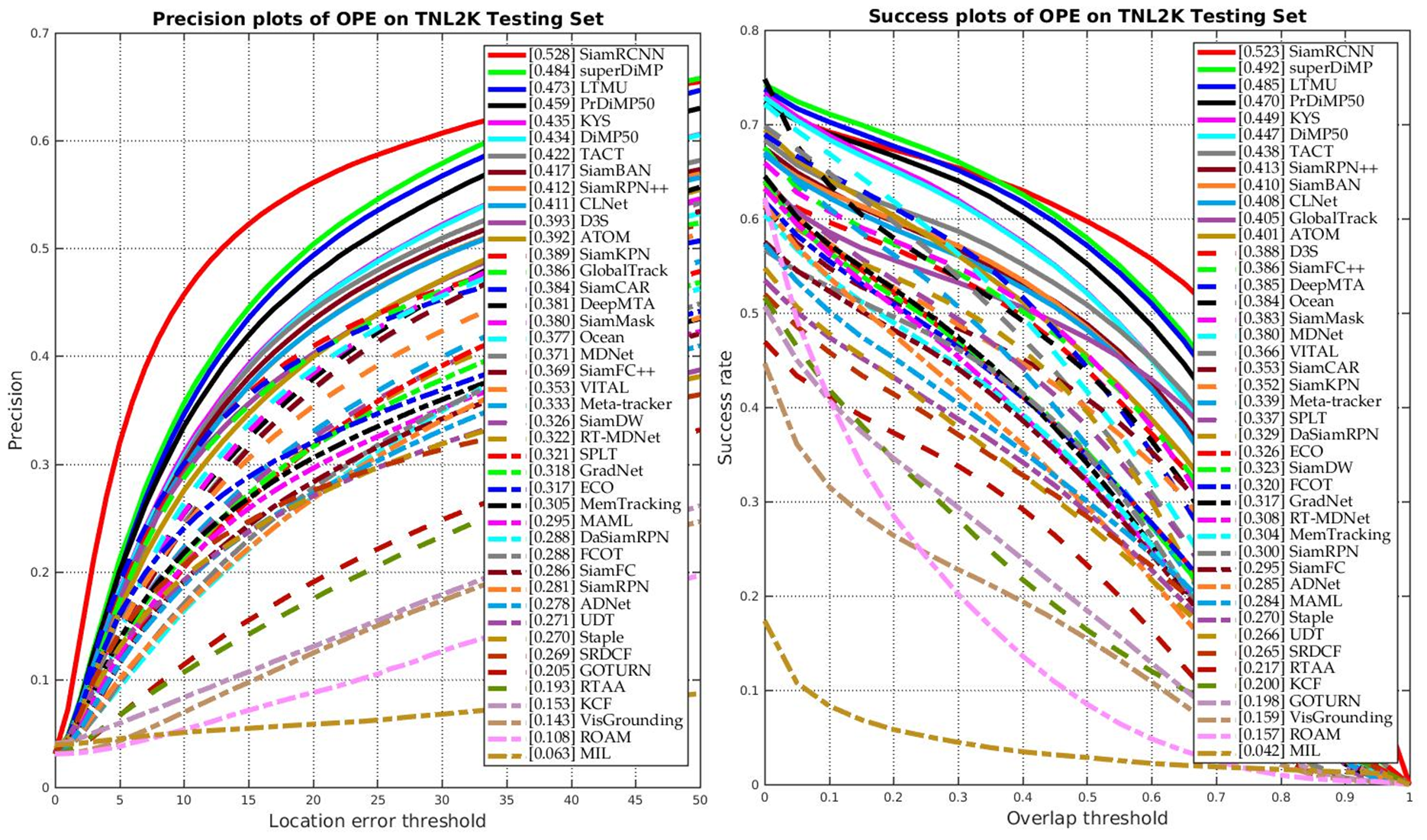}
\caption{Benchmark results of tracking-by-BBox on TNL2K dataset. Best viewed by zooming in. }
\label{benchmarkresultsBBox}
\end{figure}

\textbf{Results of Tracking by Bounding Box Only}
This setting is most widely used in existing tracking algorithms, and we provide the results of 43 representative trackers from 2015 to 2021, as shown in Fig. \ref{benchmarkresultsBBox}. These trackers contain \emph{Classification}-based, \emph{SiameseNet}-based, \emph{Correlation filter}-based, \emph{Reinforcement learning}-based, \emph{Long-term}-based and \emph{Other} trackers. More detailed introductions on these trackers can be found in our supplementary materials due to the limited space in this paper. From Fig. \ref{benchmarkresultsBBox}, we can find that SiamRCNN \cite{voigtlaender2020siamRCNN} achieves the best performance on our benchmark dataset, i.e., 0.528$|$0.523 on the precision/success plot respectively. Other trackers also attain good performance such as LTMU \cite{dai2020ltmu}, KYS \cite{Goutam2020KYS}, TACT \cite{choi2020TACT}, due to the use of joint local and global search scheme. These experiments fully demonstrate the importance of joint local and global search for visual tracking. We also find that Siamese network based trackers usually achieve better results than other trackers like multi-domain based trackers \cite{Jung_2018_ECCV, Park_2018_ECCV, Nam2015Learning}, regression based trackers \cite{held2016GOTURN}, and correlation filter based trackers \cite{Henriques2015High, bertinetto2016staple, Danelljan2016ECO}. We also notice that GlobalTrack \cite{huang2019globaltrack} which employs global search scheme only, achieves comparable performance with local search trackers \cite{zhang2020ocean, Nam2015Learning}, but worse than state-of-the-art. This may demonstrate that only global search is not enough for robust tracking. Overall, the aforementioned observations demonstrate that the structure information mining of global scene, and offline learning indeed contribute to the high-performance visual tracking.

\textbf{Results of Tracking by Joint Language and BBox}
As shown in Table \ref{Benchmarkresults}, there are five trackers designed for this setting \cite{li2017tracking, feng2020langTrackwacv, feng2019robust, yang2019grounding, wang2018describe}. Specifically, Li et al. \cite{li2017tracking} achieve 0.72$|$0.55, while Feng et al. \cite{feng2019robust, feng2020langTrackwacv} attain 0.73$|$0.67, 0.79$|$0.61 on the OTB-Lang dataset respectively. GTI \cite{yang2019grounding} combine SiamRPN++ and visual grounding module, and achieves 0.73$|$0.58, 0.47$|$0.47 on OTB-Lang and LaSOT dataset. In contrast, we can achieve 0.88$|$0.68 on the OTB-Lang, 0.55$|$0.51 on the LaSOT, 0.42$|$0.50$|$0.42 on the TNL2K (Our-II in Table \ref{Benchmarkresults}), which are significantly better than GTI \cite{yang2019grounding}, Wang et al. \cite{wang2018describe} and Feng et al. \cite{feng2020langTrackwacv}. All the experiments on three benchmark datasets validate the effectiveness and advantages of our tracker. Visualization of related tracking results can be found in Fig. \ref{trackingResults_vis}.

\begin{table}
\center
\scriptsize 
\caption{Tracking results on the OTB-Lang, LaSOT, and TNL2K dataset. $[$Prec.$|$Norm. Prec. $|$Succ. Plot$]$ are reported respectively.} \label{Benchmarkresults}
\begin{tabular}{l|c|c|c|c}
\hline \toprule [0.8 pt]
\textbf{Algorithm}    									&\textbf{Initialize}       		 &\textbf{OTB-Lang}      &\textbf{LaSOT}     			&\textbf{TNL2K}        \\
\hline 
SiamFC \cite{bertinetto2016siamfc} 			&BBox        			  	&-       						&$0.40|0.34$      						&$0.29|0.35|0.30$         			 \\	
MDNet \cite{Nam2015Learning} 					&BBox        			  	&-       						&$0.46|0.40$      						&$0.37|0.46|0.38$         			 \\	
VITAL \cite{SongYiBing_2018_CVPR} 		&BBox        			  	&-       						&$0.45|0.39$      						&$0.35|0.44|0.37$         			 \\	
GradNet \cite{Li_2019_ICCV} 					&BBox        			  	&-       						&$0.35|0.37$      						&$0.32|0.40|0.32$         			 \\	
ATOM \cite{danelljan2019atom} 					&BBox        			  	&-       						&$0.51|0.51$      						&$0.39|0.47|0.40$         			 \\	
SiamDW \cite{zhipeng2019deeper}				&BBox        			  	&-       						&$-|0.38$      							&$0.33|0.41|0.32$         			 \\		
SiamRPN++ \cite{li2018siamrpn++}			&BBox        			  	&-       						&$0.50|0.45$      						&$0.41|0.48|0.41$         			 \\		
GlobalTrack \cite{huang2019globaltrack}	&BBox        			  	&-       						&$0.53|0.52$      						&$0.39|0.46|0.41$         			 \\			
SiamBAN \cite{chen2020siamban} 				&BBox        			  	&-       						&$0.60|0.51$      						&$0.42|0.49|0.41$         			 \\	
Ocean \cite{zhang2020ocean} 						&BBox        			  	&-       						&$0.57|0.56$      						&$0.38|0.45|0.38$         			 \\
\hline \toprule [0.8 pt]
Li et al. \cite{li2017tracking}					&NL        			  	&$0.29|0.25$       		&-     						&-        			 \\
Li et al. \cite{li2017tracking}					&NL+BBox        	&$0.72|0.55$       		&-     						&-        			 \\
\hline 
Feng et al. \cite{feng2019robust} 			&NL        				&$0.56|0.54$       		&-     						&-        			 \\
Feng et al. \cite{feng2019robust} 			&NL+BBox        	&$0.73|0.67$       		&$0.56|0.50$     		&$0.27|0.34|0.25$        			 \\
\hline 
Feng et al. \cite{feng2020langTrackwacv}		&NL        				&$0.78|0.54$       		&$0.28|0.28$     		&-        			 \\
Feng et al. \cite{feng2020langTrackwacv}		&NL+BBox        	&$0.79|0.61$       		&$0.35|0.35$     		&$0.27|0.33|0.25$        			 \\
\hline 
Wang et al. \cite{wang2018describe}		&NL+BBox        	&$0.89|0.65$       		&$0.30|0.27$     		&-        			 \\ 
\hline 
GTI \cite{yang2019grounding} 				&NL+BBox       	&$0.73|0.58$       		&$0.47|0.47$     		&-        			 \\
\hline \toprule [0.8 pt]
Ours-I														&NL        				&$0.24|0.19$       	&$0.49|0.51$     			&$0.06|0.11|0.11$       \\				
Ours-II														&NL+BBox        	&$0.88|0.68$       &$0.55|0.51$			    &$0.42|0.50|0.42$       \\				
\hline \toprule [0.8 pt]
\end{tabular}
\end{table}

\subsection{Ablation Study} 

In this section, we first analyse the effectiveness of main components in our model. Then, we focus on validating the contributions of each input for AdaSwitcher. Finally, we give the parameter analysis, and attribute analysis.

\begin{figure*}[!htb]
\center
\includegraphics[width=6.3in]{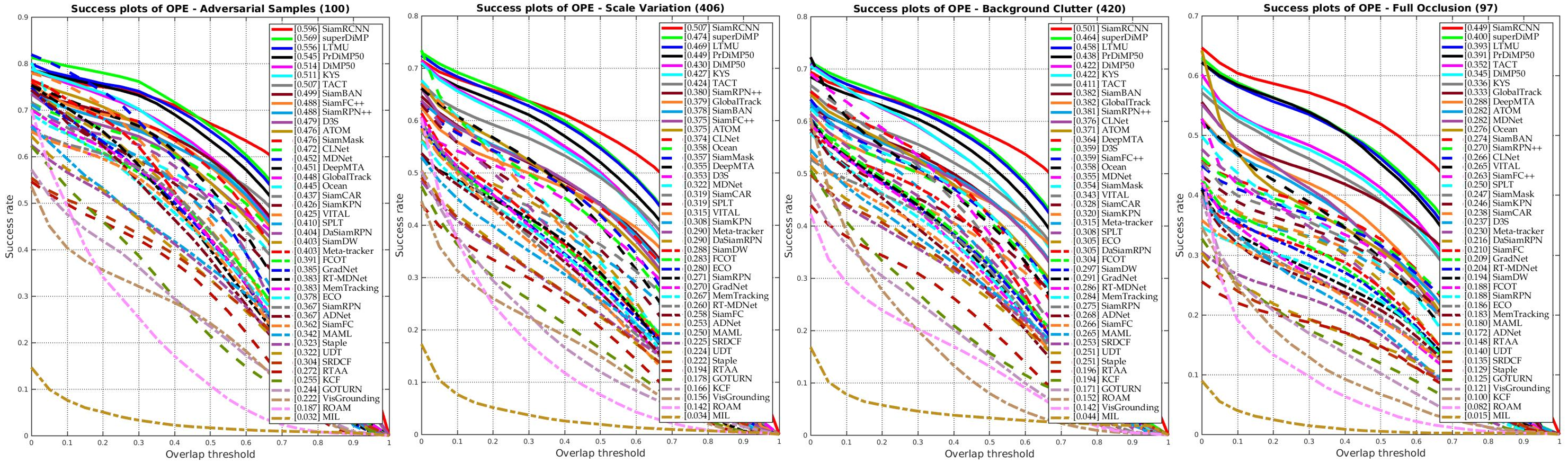}
\caption{	Tracking results under partial attributes of TNL2K dataset. Best viewed by zooming in.  }
\label{benchmarkAttresults}
\end{figure*}

\textbf{Effectiveness of AdaSwitcher } 
As shown in Table \ref{CMAnalysis_HI}, the baseline tracker SiamRPN++ \cite{li2018siamrpn++} (AlexNet version) achieves $0.344/0.353$ on the precision and success plot, respectively. When integrated with the AdaSwicher module, the performance can be improved to $0.355/0.370$. This result is also better than naive fused method (i.e. $0.347/0.362$), which fully demonstrates the effectiveness of our adaptive switch mechanism for robust tracking.

\textbf{Effectiveness of Frame Attention }
Due to different frames may contribute differently to our AdaSwitcher, we introduce the frame attention mechanism to achieve this goal. As shown in Table \ref{CMAnalysis}, with the help of frame attention, the tracking results can be improved from $0.353/0.369$ to $0.355/0.370$. This fully demonstrates the important role of frame attention in our proposed framework.

\textbf{Effectiveness of Spatial Coordinates } 
In our visual grounding module, the spatial coordinates are introduced to further improve the final results. As shown in Table \ref{CMAnalysis}, our grounding module achieves $0.143/0.159$ and $0.103/0.124$, respectively, with and without the help of spatial coordinates. This result validates the important role of spatial coordinates for visual grounding.

\begin{table}[!htp]
\center
\scriptsize 
\caption{Component analysis of our proposed tracking algorithm. AS is short for AdaSwitcher, FA denotes frame attention in AdaSwitcher, SC is spatial coordinates used in visual grounding. Naive denotes switch method based on response score only.} 
\label{CMAnalysis} 
\begin{tabular}{ccccccc|cc} 		
\hline \toprule [0.8 pt] 
Track 	     	&Ground  	&SC  					&TANet  		&Naive			&AS 	  		&FA 		 	&Results    \\
\hline 
\cmark			&    				&     				    &     			&    		 			&    		 		&    	 			&$0.344|0.353$		 	\\
\rowcolor{mygray}
					&\cmark    	&     				    &     			&    		 			&    		 		&    	 		 	&$0.103|0.124$			\\			
					&\cmark    	&\cmark     			&     			&    		 			&    		 		&    	 		 	&$0.143|0.159$			\\
\rowcolor{mygray}	
\cmark			&    				&     					&\cmark     	&\cmark    		&    		 		&    	 		 	&$0.347|0.362$			 						\\ 
\cmark			&    				&							&\cmark      &    		 			&\cmark    	&\cmark   	&$0.355|0.370$				 					\\
\rowcolor{mygray}	
\cmark			&   				&							&\cmark      &    		 			&\cmark    	&    	 		 	&$0.353|0.369$									\\
\hline \toprule [0.8 pt]
\end{tabular}
\end{table}

\textbf{Analysis on History Information }
Our AdaSwitcher takes multiple inputs for the final decision, in this section, we analyze their contributions by comparing corresponding results in Table \ref{CMAnalysis_HI}. Specifically speaking, when the BBox is discarded, we find that the performance is dropped from 0.355/0.370 to 0.350/0.365, this demonstrates that the geometric information of predicted BBox is an important clue for our tracking. Similarly, we attain worse tracking results when the resulting image (i.e. ResImg) is ignored, the results drop from 0.355/0.370 to 0.345/0.362. When all these modules removed, it attains 0.344/0.353 only on the precision and success plot. This demonstrates that this information are very important for the anomaly (or failure) detection in tracking procedure.

\begin{table}[!htp]
\center
\scriptsize   
\caption{Component analysis of history information.} \label{CMAnalysis_HI} 
\begin{tabular}{ccccc|cccc} 		
\hline \toprule [0.8 pt] 
BBox 					&Score   					&ResMap 	  		     		&ResImg						&Lang  				&Results    \\
\hline 
\cmark   				&\cmark   		 		&\cmark   	 					 &\cmark    				    &\cmark     			&$0.355|0.370$				 \\
\rowcolor{mygray}
\xmark   				&\cmark   		 		&\cmark   	 					 &\cmark    				    &\cmark     			&$0.350|0.365$				 \\
\cmark   				&\xmark   		 		&\cmark   	 					 &\cmark    				    &\cmark     			&$0.352|0.368$				 \\
\rowcolor{mygray}
\cmark   				&\cmark   		 		&\xmark   	 					 &\cmark    				    &\cmark     			 &$0.352|0.368$				 \\
\cmark   				&\cmark   		 		&\cmark   	 					 &\xmark    				    &\cmark     			 &$0.345|0.362$				\\
\rowcolor{mygray}
\cmark  				&\cmark   		 		&\cmark   	 					 &\cmark    				    &\xmark     		 	&$0.352|0.368$				\\
\xmark					&\xmark 					&\xmark				 			&\xmark			    		&\xmark				&$0.344|0.353$				 \\
\hline \toprule [0.8 pt]
\end{tabular}
\end{table}

\textbf{Parameter Analysis }
We report the tracking results with different switch thresholds in Table \ref{thresholdAnalysis}. We can find that the performance is better when switch threshold is set as 0.7.

\begin{table}[!htp]
\center
\scriptsize 
\caption{Results with different switch threshold.}  
\label{thresholdAnalysis} 
\begin{tabular}{c|cccccccc}
\hline \toprule [0.8 pt] 
Parameter 			&0.5			&0.6  			&0.7  			&0.8				&0.9 	  		&1.0 		 		&1.2    			\\
\hline 
Prec. Plot		&0.350		&0.351  		&0.355  		&0.353			&0.349 	  	&0.352 		 	&0.272    		\\
\hline 
Succ. Plot		&0.367		&0.367  		&0.370  		&0.369			&0.368 	  	&0.368 		 	&0.301    		\\
\hline \toprule [0.8 pt]
\end{tabular}
\end{table}

\begin{figure}[!htb]
\center
\includegraphics[width=3.3in]{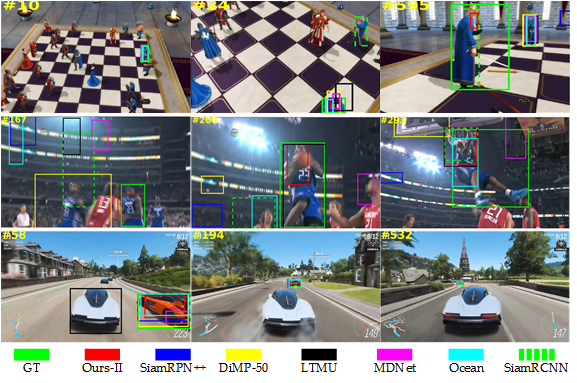}
\caption{Visualization of tracking results on TNL2K dataset. }
\label{trackingResults_vis}
\end{figure}

\textbf{Attribute Analysis } 
Evaluation under each challenging factors is one of the most important metrics in visual tracking community. In this benchmark, we also report results of evaluated trackers under all the defined 17 attributes. However, due to limited space in this paper, we select 4 attributes, i.e., \emph{Adversarial Samples}, \emph{Scale Variation}, \emph{Background Clutter}, and \emph{Full Occlusion}, to demonstrate the ability of resistance of these trackers to these challenges. As shown in Fig. \ref{benchmarkAttresults}, we can find that SiamRCNN \cite{voigtlaender2020siamRCNN} achieves the best performance which are much better than the second and third ones, i.e., DiMP \cite{bhat2019DiMP} and LTMU \cite{dai2020ltmu} respectively. Interestingly, it is easy to find that the RTAA \cite{jia2020TrackAttack} which is designed for adversarial attack achieves worse results on the challenging factor Adversarial Samples, even compared with their baseline DaSiamRPN \cite{zhu2018distractor}. This demonstrates that the detection of adversarial samples is important for high performance tracking. More experimental results on the attribute analysis can be found in our supplementary materials.

\section{Conclusion and Future Works} 
In this paper, we revisit the tracking by natural language, and propose a large-scale benchmark for this task. Specially, a large-scale dataset that contains 2,000 video sequences is proposed, named TNL2K. This dataset is densely annotated with bounding box and natural language description of target object. To construct a sound benchmark, we propose an adaptive switch based tracking algorithm as the baseline approach, i.e., the AdaSwicher, and also test current trackers according to following settings: tracking by natural language only, tracking by bbox, and tracking by joint bbox and language. We believe our benchmark will be greatly boost related researches on the natural language guided tracking. In our future works, we will consider to further extend this benchmark by introducing more videos and baseline trackers. Besides, we will focus on improving the visual grounding module to achieve high performance language initialized tracking.

\noindent \scriptsize{\textbf{Acknowledgement: } This work is jointly supported by Key-Area Research and Development Program of Guangdong Province 2019B010155002, Postdoctoral Innovative Talent Support Program BX20200174, China Postdoctoral Science Foundation Funded Project 2020M682828, National Natural Science Foundation of China (61976002, 61825101), National Key Research and Development Program of China 2020AAA0106800.}

\small{ 
\bibliographystyle{ieee_fullname}
\bibliography{reference}
}

\clearpage

%
  
\appendix

\section{The TNL2K Benchmark}

\subsection{Motivation and Protocols} 
\begin{figure*} 
\center
\includegraphics[width=7in]{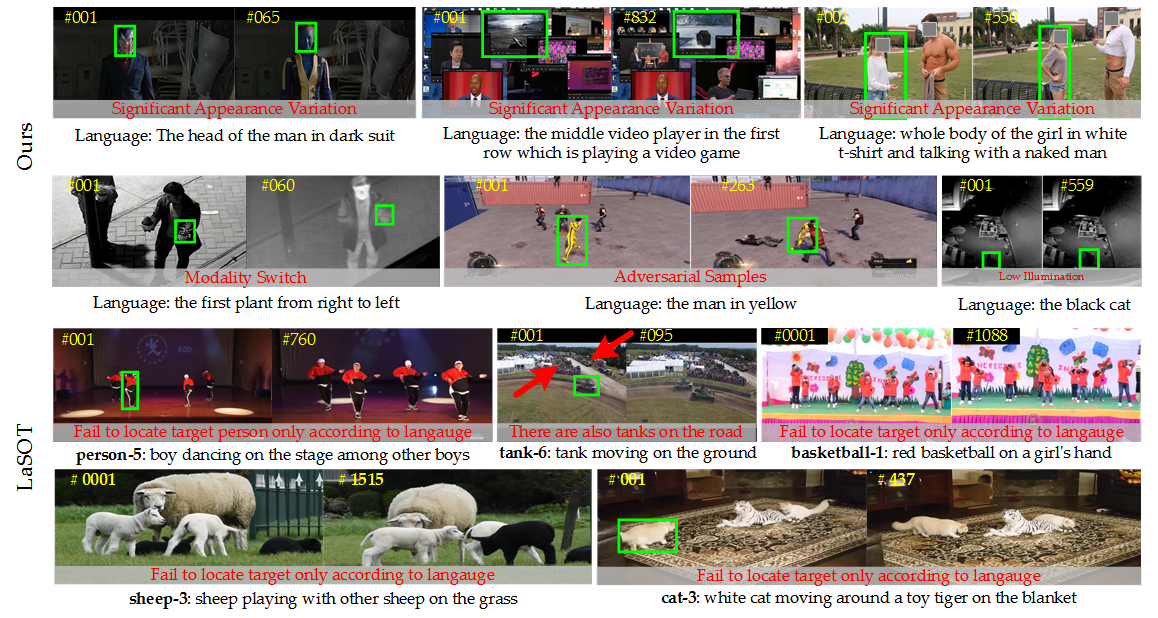}
\caption{Comparison between our proposed TNL2K dataset and existing LaSOT dataset. Best viewed by zooming in.} 
\label{compare_tnl2k_lasot}
\end{figure*}

\textbf{Motivation}: Directly extending existing datasets like GOT-10k \cite{huang2019got10k} is an intuitive and good idea for this task, but GOT-10k contains few videos with special properties as mentioned in Fig. 1 in our paper. Also, its videos are all short-term which can't reflect performance gain of re-detection with language. As for LaSOT \cite{fan2019lasot}, many of its language annotations can not point out target object clearly, as shown in Fig. \ref{compare_tnl2k_lasot}. Thus, LaSOT is not suitable for tracking-by-language only. Similar views can also be found in GTI \cite{yang2019grounding}. Therefore, we build the TNL2K (from video collection, dense bbox and language annotation, to diverse baseline construction) to better reflect the characteristics (see below) of tracking by natural language. The target of this work is not to construct the largest tracking dataset, but to build the first benchmark specifically designed for tracking-by-language task. Compared with GOT-10k and LaSOT, the data collection of TNL2K is a compromise between length and quantity.

\textbf{Protocol}: When collecting the videos, we attempt to search the target object is \emph{severely occluded in the first frame}, with \emph{significant appearance variation} (e.g., cloth changing for human),  \emph{can only be located with reasoning}, which correspond to Fig. \ref{motivation} in our paper. Also, we collect videos from other thermal tracking datasets and annotate language descriptions only to check the robustness to certain challenging factors like domain adaptation, modality switch, etc.

\subsection{Why add Attribute Modality Switch (MS) ?} 
In the proposed TNL2K dataset, we design a new attribute termed Modality Switch (MS) for object tracking. This is mainly motivated by the fact that the RGB cameras work well in the daytime but nearly ineffective at night, meanwhile, the thermal cameras work well in the night time. If we track a target for an extremely long-term (e.g., several days or weeks), collaboration between RGB and thermal cameras are needed. Therefore, the connections between the two modalities need to be set up. Similar views can be found in cross-modality person re-identification \cite{wu2017rgbinfraredreid, wu2020rgbirreid}. There are still no works on object tracking try to build such connections and they usually study these two cameras separately (i.e., RGB tracking \cite{Wu2013Online, fan2019lasot, Wang_2018_CVPR}, Thermal Tracking \cite{liu2019ptbtracking}) or in an integrated approach (i.e., RGB-T tracking \cite{li2019rgbt234}). In this work, we propose the modality switch and attempt to encourage researches on such cross-modality object tracking.

\subsection{Highlights of TNL2K Dataset}

Generally speaking, our proposed benchmark TNL2K have the following features as shown in Table \ref{benchmarkList}:  
\begin{itemize}

\item \textbf{TNL2K is the first benchmark specifically designed for tracking-by-natural language.} Different from regular tracking benchmarks like OTB, GOT10k, and TrackingNet, we provide both language annotation and dense bounding box annotation for each video sequence which will be a good platform for natural language-related tracking. Different from the recently released long-term tracking dataset LaSOT which also provides language annotation, their annotation only describes the attribute of target object, but ignores the spatial position. Therefore, this benchmark can be only used for the task of \emph{tracking by joint language and bbox}. Our language annotations not only embody the attribute, category, shape, properties, and structural relationship with other objects, therefore, our dataset can also be used for the task of \emph{tracking by natural language only}. Some video sequences and corresponding annotations are provided in Figure \ref{compare_tnl2k_lasot} to give an intuitive understanding of the difference between our TNL2K and LaSOT. 

\item \textbf{TNL2K is the first benchmark to provides videos with actively introduced adversarial samples} which will be beneficial for the development of adversarial training for tracking. 

\item \textbf{TNL2K is the first benchmark to provides videos with significant appearance variation}, such as \emph{cloth/face changing}. We believe our benchmark will greatly boost related research on abrupt appearance variation based tracking.  

\item \textbf{TNL2K provides a heterogeneous dataset} that contains RGB video, Thermal video \footnote{There are 518 videos totally borrowed from existing RGB-T dataset \cite{li2019rgbt234} and infrared tracking dataset \cite{liu2019ptbtracking}.}, Cartoon, and Synthetic data (i.e., videos from games). It can be used for the study of domain adaptation, e.g., train the tracker on RGB data and test it on Thermal videos. 

\item \textbf{TNL2K provides three kinds of baseline methods for future works to compare}, including Tracking-by-BBox, Tracking-by-Language, Tracking-by-Joint-BBox-Language.

\end{itemize}

\section{The Proposed Method}

\subsection{	YOLO Loss and BCE Loss Functions} 
In the training phase, we use the YOLO loss function for the optimization of the visual grounding module by following \cite{yang2019fastgrounding}. This loss is first proposed in YOLOv3 \cite{redmon2018yolov3}  which attempt to predict the five quantities of each anchor box by shifting its \emph{center, width, height,} and the \emph{confidence} on this shifted box. To better use it for visual grounding, the following two changes are modified by Yang et al.: 1). recalibrate its anchor boxes; 2). change its sigmoid layer to a softmax function. Due to the object detection is designed for output multiple locations, while visual grounding only needs to predict one bbox which best fit the language description. Therefore, the sigmoid function in YOLOv3 is replaced by softmax function. The cross-entropy is used for the measurement of confidence scores, and the regions with maximum IoU with ground truth are labeled as 1, other regions are set as 0. More details can be found in \cite{redmon2018yolov3, yang2019fastgrounding}. For the training of TANet, we adopt Binary Cross-Entropy (BCE) loss to measure the distance between the ground truth mask and the prediction.

\subsection{Details of Evaluated Trackers } 
In this section, we provide the details of evaluated BBox-based trackers on our TNL2K dataset. As shown in Table \ref{SummaryTrackers}, the publication, feature representation, update or not, need pre-train or not, search scheme, tracking efficiency, and results (Precision Plot and Success Plot) on the TNL2K are all reported. These tracking algorithms are ranked according to the results. 

\begin{table*}[!htp]
\center
\scriptsize  
\caption{Summary of evaluated trackers on TNL2K dataset.}  
\label{SummaryTrackers}  
\begin{tabular}{rrrrccrrccc} 		
\hline \toprule [0.8 pt] 
\textbf{Index} 		&\textbf{Tracker}  	&\textbf{Publication}  &\textbf{Feature}      		&\textbf{Update}			&\textbf{Pre-train}				&\textbf{Search Scheme} 		&\textbf{FPS} 		 &\textbf{Results}    \\
\hline 
001			&SiamRCNN	 \cite{voigtlaender2020siamRCNN}		&CVPR-2020    	&ResNet-101     	&\xmark    		 	&\cmark    	&Local + Global	&5@GPU    		&$0.528|0.523$\\
002			&SuperDiMP    \cite{bhat2019DiMP}							&ICCV-2019    	&ResNet-50     	&\cmark     			&\cmark    	&Local    		 		&40@GPU    	&$0.484|0.492$		 	\\
003			&LTMU 			 \cite{dai2020ltmu}								&CVPR-2020    	&ResNet-50     	&\cmark     			&\cmark    	&Local    		 		&13@GPU    	&$0.473|0.485$		 	\\
004			&PrDiMP50 	 \cite{danelljan2020PRDiMP}				&CVPR-2020    	&ResNet-50     	&\cmark     			&\cmark    	&Local    		 		&30@GPU    	&$0.459|0.470$		 	\\
005			&KYS			 	 \cite{Goutam2020KYS}						&ECCV-2020    	&ResNet-50     	&\cmark     			&\cmark    	&Local + Global  &20@GPU    	&$0.435|0.449$		 	\\
006			&DiMP50	 		 \cite{bhat2019DiMP}							&ICCV-2019    	&ResNet-50     	&\cmark     			&\cmark    	&Local    		 		&40@GPU    	&$0.434|0.447$		 	\\
007			&TACT		 	 \cite{choi2020TACT}							&ACCV-2020    	&ResNet-50     	&\xmark    			&\cmark    	&Local + Global  &42@GPU    	&$0.422|0.438$		 	\\
008			&SiamBAN		 \cite{chen2020siamban}						&CVPR-2020    	&ResNet-50     	&\xmark     			&\cmark    	&Local    		 		&40@GPU    	&$0.417|0.410$		 	\\
009			&SiamRPN++	 \cite{li2018siamrpn++}						&CVPR-2019    	&ResNet-50     	&\xmark     			&\cmark    	&Local    		 		&35@GPU    	&$0.412|0.413$		 	\\
010			&CLNet		 	 \cite{dong2020clnet}							&ECCV-2020    	&ResNet-50     	&\xmark     			&\cmark    	&Local    		 		&45@GPU    	&$0.411|0.408$		 	\\
011			&D3S			 	 \cite{lukezic2020d3s}							&CVPR-2020    	&ResNet-50     	&\xmark     			&\cmark    	&Local    		 		&25@GPU    	&$0.393|0.388$		 	\\
012			&ATOM		 	 \cite{danelljan2019atom}					&CVPR-2019    	&ResNet-50     	&\xmark     			&\cmark    	&Local    		 		&30@GPU    	&$0.392|0.401$		 	\\
013			&SiamKPN		 \cite{li2020siamKPN}							&arXiv-2020    	&ResNet-50     	&\xmark     			&\cmark    	&Local    		 		&24@GPU    	&$0.389|0.352$		 	\\
014			&GlobalTrack	 \cite{huang2019globaltrack}				&AAAI-2020    	&ResNet-50     	&\xmark     			&\cmark    	&Global    		 	&6@GPU    	 	&$0.386|0.405$		 	\\
015			&SiamCAR		 \cite{guo2020siamcar}						&CVPR-2020    	&ResNet-50     	&\xmark     			&\cmark    	&Local    		 		&52@GPU    	&$0.384|0.353$		 	\\
016			&DeepMTA		\cite{deepMTA}									&TCSVT-2021    	&ResNet-50     	&\cmark     			&\cmark    	&Local + Global    	&12@CPU   				&$0.381|0.385$		 	\\
017			&SiamMask	 	 \cite{wang2019fast}							&CVPR-2019    	&ResNet-50     	&\xmark     			&\cmark    	&Local    		 		&55@GPU    	 			&$0.380|0.383$		 	\\
018			&Ocean		 	  \cite{zhang2020ocean}						&ECCV-2020    	&ResNet-50     	&\xmark     			&\cmark    	&Local    		 		&58@GPU    	 			&$0.377|0.384$		 	\\
019			&MDNet		 	  \cite{Nam2015Learning}					&CVPR-2016    	&CNN-3     			&\cmark     			&\cmark    	&Local    		 		&1@GPU    	 				&$0.371|0.384$		 	\\
020			&SiamFC++	  	  \cite{xu2020siamfc++}						&AAAI-2020    	&GoolgeNet     	&\cmark     			&\cmark    	&Local    		 		&90@GPU    	 			&$0.369|0.386$		 	\\
021			&VITAL			  \cite{SongYiBing_2018_CVPR}			&CVPR-2018    	&CNN-3     			&\cmark     			&\cmark    	&Local    		 		&1.5@GPU    	 			&$0.353|0.366$		 	\\
022			&Meta-Tracker \cite{Park_2018_ECCV}						&ECCV-2018   	&CNN-3     			&\cmark     			&\cmark    	&Local    		 		&1@GPU    	 				&$0.333|0.339$		 	\\
023			&SiamDW		 \cite{zhipeng2019deeper}					&CVPR-2019    	&Res22W     		&\xmark     			&\cmark    	&Local    		 		&150@GPU   	 			&$0.326|0.323$		 	\\
024			&RT-MDNet	 	\cite{Jung_2018_ECCV}						&ECCV-2018    	&CNN-3     			&\cmark     			&\cmark    	&Local    		 		&46@GPU    	 			&$0.322|0.308$		 	\\
025			&SPLT			 	\cite{yan2019skimming}						&ICCV-2019    	&ResNet-50     	&\xmark     			&\cmark    	&Local + Global  &25@GPU    	 			&$0.321|0.337$		 	\\
026			&GradNet		 \cite{Li_2019_ICCV}							&ICCV-2019    	&CNN-5     			&\cmark     			&\cmark    	&Local    		 		&80@GPU   				&$0.318|0.317$		 	\\
027			&ECO				 \cite{Danelljan2016ECO}					&CVPR-2017    	&VGG     			&\cmark     			&\xmark    	&Local    		 		&8@CPU    	 				&$0.317|0.326$		 	\\
028			&MemTracking \cite{Yang_2018_ECCV}					&ECCV-2018    	&CNN-5     			&\cmark     			&\cmark    	&Local    		 		&50@GPU    	 			&$0.305|0.304$		 	\\
029			&MAML			 \cite{wang2020MAML}						&CVPR-2020    	&ResNet-50     	&\xmark     			&\cmark    	&Local    		 		&40@GPU    	 			&$0.295|0.284$		 	\\
030			&DaSiamRPN	 \cite{zhu2018distractor}						&ECCV-2018    	&ResNet-50     	&\xmark     			&\cmark    	&Local    		 		&110@GPU    	 			&$0.288|0.329$		 	\\
031			&FCOT			 \cite{cui2020FCOT}							&arXiv-2020    	&ResNet-50     	&\xmark     			&\cmark    	&Local    		 		&45@GPU    	 			&$0.288|0.320$		 	\\
032			&SiamFC		 	 \cite{bertinetto2016siamfc}					&ECCVW-2016   &CNN-5     			&\xmark     			&\cmark    	&Local    		 		&58@GPU    	 			&$0.286|0.295$		 	\\
033			&SiamRPN		 \cite{li2018siamRPN}							&CVPR-2018    	&ResNet-50     	&\xmark     			&\cmark    	&Local    		 		&160@GPU    	 			&$0.281|0.300$		 	\\
034			&ADNet			 \cite{Yun2017ADNet}							&CVPR-2017    	&CNN-3     			&\cmark     			&\cmark    	&Local    		 		&3@GPU    	 				&$0.278|0.285$		 	\\
035			&UDT				 \cite{wang2019UDT}							&CVPR-2019    	&CNN-5     			&\xmark     			&\cmark   	&Local    		 		&70@GPU    	 			&$0.271|0.266$		 	\\
036			&Staple			 \cite{bertinetto2016staple}					&CVPR-2016    	&HOG     			&\cmark     			&\xmark    	&Local    		 		&80@CPU    	 			&$0.270|0.270$		 	\\
037			&SRDCF		 	 \cite{danelljan2015srdcf}					&ICCV-2015    	&HOG     			&\cmark     			&\xmark    	&Local    		 		&6@CPU    	 				&$0.269|0.265$		 	\\
038			&GOTURN		 \cite{held2016GOTURN}					&ECCV-2016    	&CaffeNet-5     	&\xmark     			&\cmark    	&Local    		 		&100@GPU    	 			&$0.205|0.198$		 	\\
039			&RTAA			 \cite{jia2020TrackAttack}					&ECCV-2018    	&ResNet-50     	&\xmark     			&\cmark    	&Local    		 		&2.2@GPU    	 			&$0.193|0.217$		 	\\
040			&KCF				 \cite{Henriques2015High}					&TPAMI-2015    	&HOG     			&\cmark     			&\xmark    	&Local    		 		&172@CPU    	 			&$0.153|0.200$		 	\\
041			&VisGround	 \cite{yang2019fastgrounding}				&ICCV-2019    	&DarkNet-53     	&\xmark     			&\cmark    	&Global    		 	&147@GPU    	 			&$0.143|0.159$		\\  
042			&ROAM			 \cite{yang2020roam}							&CVPR-2020    	&ResNet-50     	&\xmark     			&\cmark    	&Local    		 		&13@GPU    	 			&$0.108|0.157$		 	\\
043			&MIL				 \cite{babenko2009MIL}						&CVPR-2009    	&HOG     			&\cmark     			&\xmark    	&Local    		 		&25@CPU   				&$0.063|0.042$		 	\\
\hline \toprule [0.8 pt]
\end{tabular}
\end{table*}

\subsection{Introduction to TANet} 

Inspired by \cite{wang2019GANTrack, wang2018describe, deepMTA}, we introduce the TANet for the global search to replace the Grounding module \cite{yang2019fastgrounding} in the setting of \emph{tracking-by-joint language and BBox}, termed Ours-II. Generally speaking, the TANet is inspired by semantic segmentation, which takes the target object and video frames as input and output an attention map using a decoder network. The estimated attention maps can highlight the possible search regions from a global view. Therefore, it can be seen as a kind of global search scheme and can be integrated with the baseline tracker and our proposed AdaSwitcher module for robust and accurate tracking. Our experimental results also demonstrate that we can attain good performance on three used datasets, i.e., the OTB-Lang \cite{li2017tracking}, LaSOT \cite{fan2019lasot}, and TNL2K. This will be a strong baseline method for future works to compare on the language guided visual tracking. The implementation of our all networks will be released for other researchers to follow.

\section{Experimental Results}  

%
%
%

\subsection{Attribute Analysis } 
\begin{figure*}[!htb]
\center
\includegraphics[width=7in]{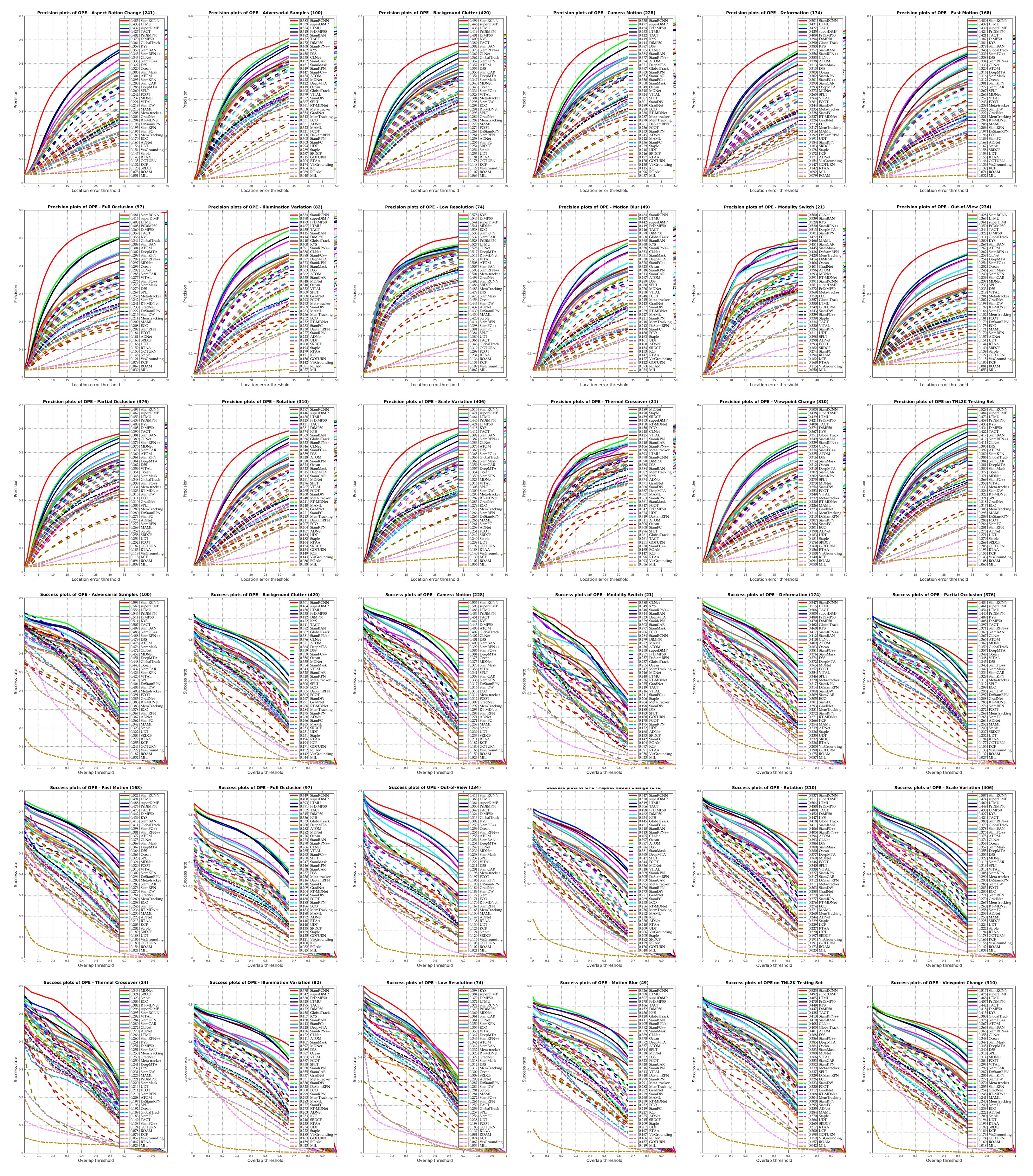}
\caption{Tracking results under each challenging factors on TNL2K dataset (Tracking-by-BBox). Best viewed by zooming in.}
\label{benchmarkAttresults_ALL}
\end{figure*}

As shown in Figure \ref{benchmarkAttresults_ALL}, we provide experimental results of all the defined 17 attributes of our TNL2K dataset. Generally speaking, we can find that the SiamRCNN  \cite{voigtlaender2020siamRCNN} achieves the best performance on most of the attributes, like \emph{Scale Variation, Rotation, Background Clutter, Partial Occlusion, Adversarial Samples, Deformation, Fast Motion, Out-of-view, Motion Blur, Aspect Ration Change, Illumination Variation, Camera Motion}, and \emph{Viewpoint Change}. Meanwhile, the SuperDiMP \cite{bhat2019DiMP}, LTMU \cite{dai2020ltmu}, PrDiMP \cite{danelljan2020PRDiMP} and KYS  \cite{Goutam2020KYS} also attains good performance on these attributes, and the KYS also achieves top-1 results on the \emph{Low Resolution}. These results all demonstrate the strong performance of Siamese network based trackers with the help of pre-training and joint local and global search scheme. Interestingly, we can also find that on the attribute \emph{Thermal Crossover} which are all thermal videos, the MDNet \cite{Nam2015Learning} which is an online learned tracker attain the best results. Even the Staple and SRDCF are better than most of the other Siamese trackers, such as SiamKPN, SiamCAR, SiamRPN++, SiamRCNN, KYS, etc. The huge contrast demonstrates that online learning is very important for the tracker which is trained on one domain and tested on another domain (for example, the tracker trained on RGB videos and tested on Thermal videos).

\subsection{Efficiency Analysis} 
In this work, two baseline methods are proposed for the \emph{natural language initialized tracking} (Our-I) and \emph{natural language guided tracking} (Our-II). 
For Our-I, the overall running efficiency is 24.39 FPS on the OTB-Lang, tested on a laptop with Intel Core I7, RTX2070.  
For Our-II, the overall efficiency on the OTB-Lang is 12.44 FPS.

\subsection{More Visualization } 

\begin{figure*}[!htb]
\center
\includegraphics[width=5.5in]{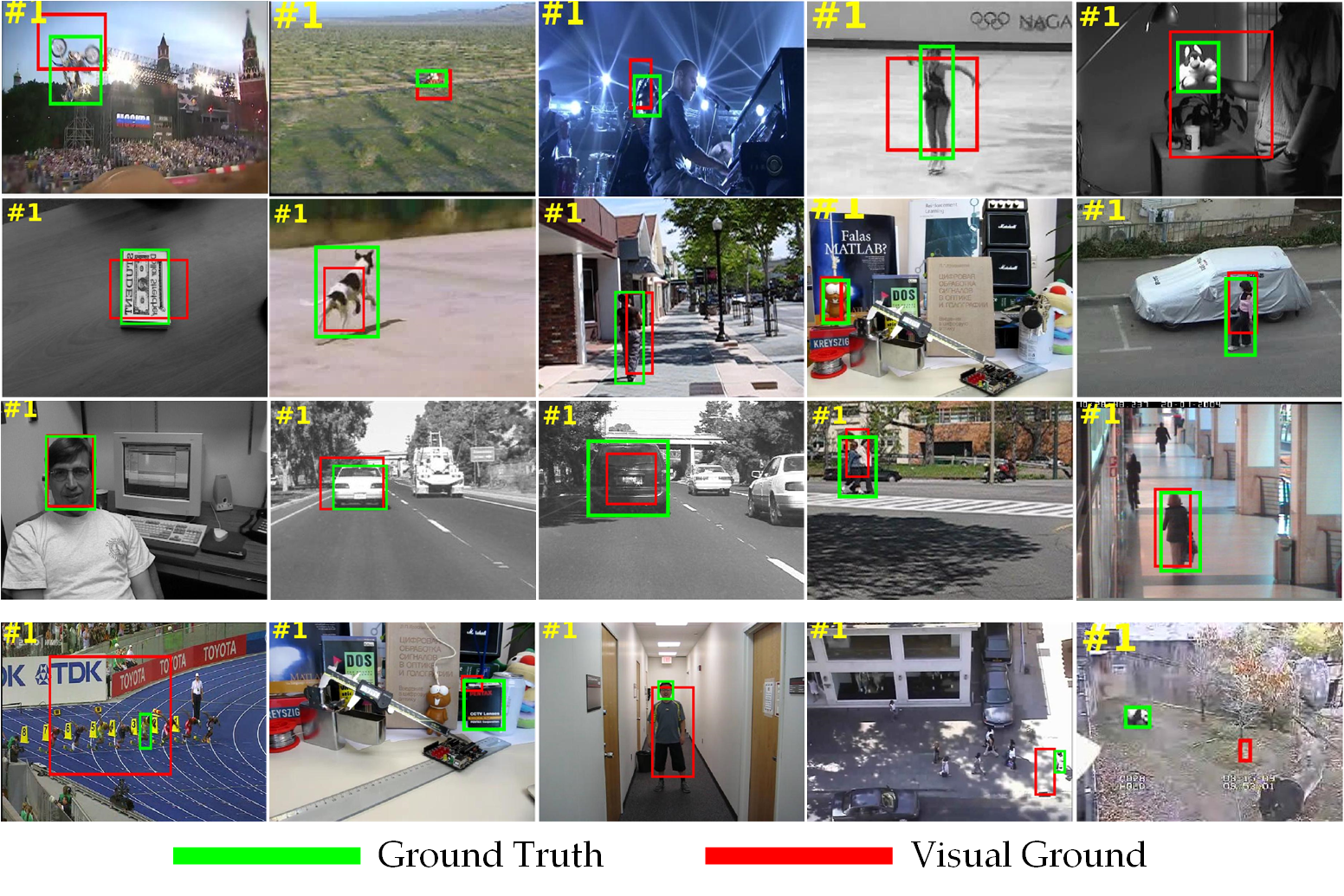}
\caption{Results of the first frame of visual grounding module. }
\label{otb99lang_vis}
\end{figure*}

In this section, more visualization on the tracking results is given to better understand our proposed method. As shown in Figure \ref{otb99lang_vis}, 20 video sequences from OTB-Lang are selected to demonstrate the results of the visual grounding module. From the first three rows, we can find that the grounding module can locate the target object accurately when the background is relatively clean. Also, it works well in some challenge videos, like \emph{car}, and \emph{human head}. For the fourth row, the grounding is not accurate enough for tracking, including the central location and scale. We can find that the performance of visual grounding is needed to be further improved for more accurate tracking. More experimental results of our proposed baseline and other trackers can be found in Figure \ref{trackingResults_vis}.

\begin{figure*}[!htb]
\center
\includegraphics[width=5.5in]{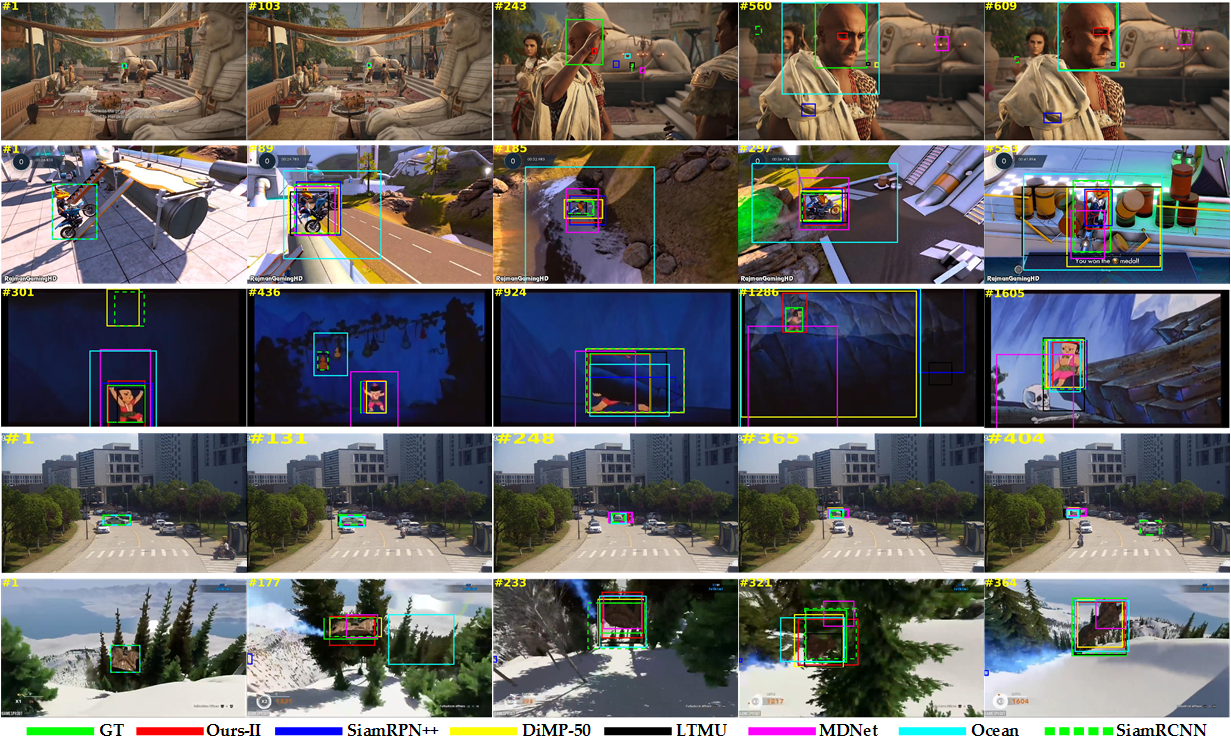}
\caption{Tracking results of our method and other state-of-the-art tracking algorithms. } 
\label{trackingResults_vis}
\end{figure*}

\end{document}